\documentclass[10pt,twocolumn,letterpaper]{article}

\usepackage{cvpr}              

\usepackage{graphicx}
\usepackage{amsmath}
\usepackage{amssymb}
\usepackage{booktabs}
\usepackage{times}
\usepackage{epsfig}
\usepackage{float}
\usepackage{multirow}
\usepackage{enumitem}
\usepackage{graphicx, amsmath, amssymb, caption, subcaption, multirow, overpic, textpos}
\usepackage[table,dvipsnames]{xcolor}
\usepackage[british, english, american]{babel}
\definecolor{citecolor}{HTML}{0071BC}
\definecolor{linkcolor}{HTML}{ED1C24}
\usepackage[colorlinks,linkcolor=blue]{hyperref}
\def\cvprPaperID{**}
\def\confName{****}
\def\confYear{****}

\newlength\savewidth
\newcommand{\tablestyle}[2]{\setlength{\tabcolsep}{#1}\renewcommand{\arraystretch}{#2}\centering\footnotesize}
\renewcommand{\paragraph}[1]{\vspace{1.25mm}\noindent\textbf{#1}}

\newcolumntype{x}[1]{>{\centering\arraybackslash}p{#1pt}}
\newcolumntype{y}[1]{>{\raggedright\arraybackslash}p{#1pt}}
\newcolumntype{z}[1]{>{\raggedleft\arraybackslash}p{#1pt}}

\newcommand{\app}{\raise.17ex\hbox{$\scriptstyle\sim$}}

\definecolor{deemph}{gray}{0.6}

\definecolor{baselinecolor}{gray}{.9}

\begin{document}
	\title{Bridging Video-text Retrieval with \emph{Multiple Choice Questions}}
	
	\author{Yuying Ge$^1$ \quad
		Yixiao Ge$^2$ \quad
		Xihui Liu$^5$ \quad
		Dian Li$^4$ \quad
		Ying Shan$^2$ \quad
		Xiaohu Qie$^3$ \quad
		Ping Luo$^1$ \quad\\
		{$^1$The University of Hong Kong} \quad 
		{$^2$ARC Lab, $^3$Tencent PCG}  \quad\\
		{$^4$Content Understanding Center, $^3$Tencent PCG}  \quad
		{$^5$UC Berkeley}\\
		\tt\small{yuyingge@hku.hk \quad \{yixiaoge, goodli, yingsshan, tigerqie\}@tencent.com}\\
		\tt\small{xihui.liu@berkeley.edu \quad pluo@cs.hku.hk}\\
		\vspace{5pt}
		{\small \url{https://github.com/TencentARC/MCQ}}
	}
	
		\begin{figure}[htb]
	\twocolumn[{
		\maketitle
		\vspace{-25pt}
		\begin{center}
			\centering
			\vspace{-6pt}
			\includegraphics[width=0.9\textwidth]{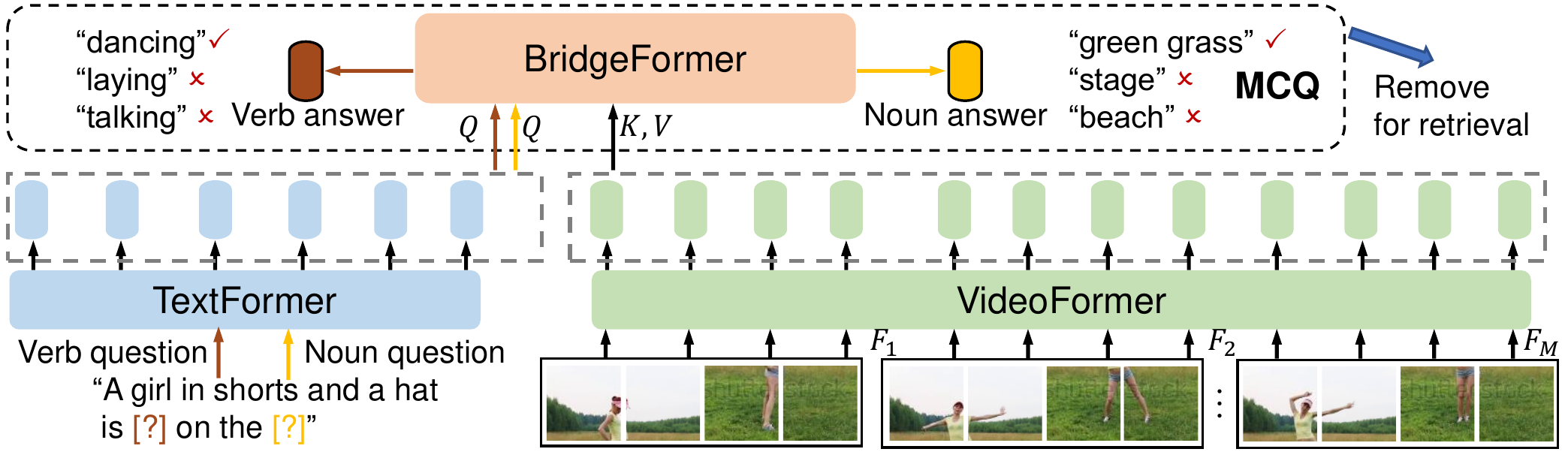}
			\vspace{-4mm}
		\end{center}
		\caption{Overview of our novel pretext task, Multiple Choice Questions (MCQ), for video-text pre-training. 
			MCQ is performed using a newly-proposed parametric module BridgeFormer, which associates all-level local features (intermediate tokens) from VideoFormer and TextFormer to answer multiple choice questions in the form of contrastive learning. 
			Given that nouns and verbs carry informative local objects and object motions,
			we construct \textbf{a noun question} (in yellow) and \textbf{a verb question} (in red) by erasing the corresponding phrase from the sentence.
			The BridgeFormer is trained to select the correct erased phrase via visual reasoning with intermediate tokens from VideoFormer, given the questions' intermediate tokens from TextFormer.
			The noun and verb questions promote VideoFormer to capture detailed spatial content and temporal information. 
			The semantic associations between video-text intermediate tokens are also enhanced via the proxy task of questions and answers.
			Note that BridgeFormer is \textbf{removed} for downstream retrieval. 
		}
		\label{fig:framework}
		\vspace{15pt}
	}	]
\end{figure}
\maketitle
\vspace{-25pt}
\begin{abstract}
\vspace{-5pt}
	Pre-training a model to learn transferable video-text representation for retrieval has attracted a lot of attention in recent years.
	Previous dominant works mainly adopt two separate encoders for efficient retrieval, but ignore local associations between videos and texts. Another line of research uses a joint encoder to interact video with texts, but results in low efficiency since each text-video pair needs to be fed into the model. 
	In this work, we enable fine-grained video-text interactions while maintaining high efficiency for retrieval via a novel pretext task, dubbed as Multiple Choice Questions (MCQ), where a parametric module BridgeFormer is trained to answer the ``questions'' constructed by the text features via resorting to the video features.
    Specifically, we exploit the rich semantics of text (i.e., nouns and verbs) to build questions, with which the video encoder can be trained to capture more regional content and temporal dynamics.
    In the form of questions and answers, the semantic associations between local video-text features can be properly established.
    BridgeFormer is able to be removed for downstream retrieval, rendering an efficient and flexible model with only two encoders.
	Our method outperforms state-of-the-art methods on the popular text-to-video retrieval task in five datasets with different experimental setups (i.e., zero-shot and fine-tune), including HowTo100M (one million videos). We further conduct zero-shot action recognition, which can be cast as video-to-text retrieval, and our approach also significantly surpasses its counterparts. %
	As an additional benefit, our method achieves competitive results with much shorter pre-training videos on single-modality downstream tasks, e.g., action recognition with linear evaluation. 
\end{abstract}

\vspace{-10pt}
\section{Introduction}
\label{sec:intro}
\begin{figure}
	\centering
	\includegraphics[width=0.9\linewidth]{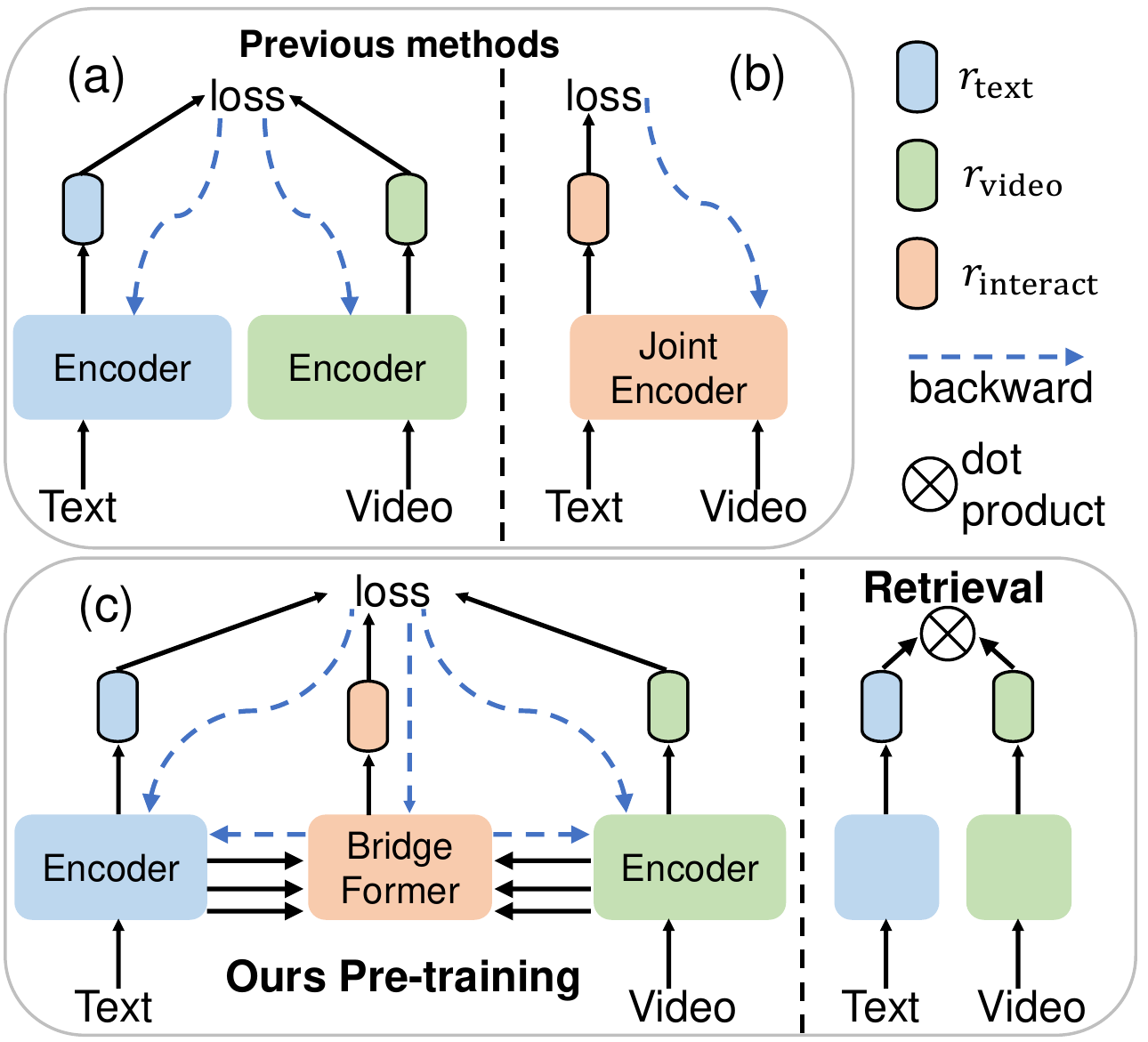}
	\caption{Comparison between existing paradigms and ours for video-text pre-training. Previous dominant methods 
		either \textbf{(a)} adopt two separate encoders to contrast video-level and sentence-level representations, ignoring local associations between videos and texts, or \textbf{(b)} use a joint encoder to interact fine-grained features of videos and texts through concatenating them as inputs, resulting in low efficiency for retrieval. \textbf{(c)} We propose a novel pretext task that uses a BridgeFormer to promote local feature learning and fine-grained video-text associations. For downstream retrieval task, the proxy BridgeFormer is \textbf{removed}.}
	\label{fig:compare}
	\vspace{-4mm}	
\end{figure}

Pre-training a model to learn transferable representations for video-text retrieval requires the understanding of video concepts, text semantics, and the relationships between videos and texts.
Existing works for video-text pre-training can be divided into two main categories. 
``Dual-encoder'' methods \cite{frozen, MIL-NCE, videoclip, taco, coot,avlnet,support, multi,expert} (see Fig.~\ref{fig:compare} (a)) adopt two separate encoders to contrast video-level and sentence-level representations respectively, ignoring the detailed local information within each modality and the associations between modalities.
``Joint-encoder'' methods \cite{clipbert,videobert,actbert,hero, univl, vlm} (see Fig.~\ref{fig:compare} (b)) concatenate texts and videos as inputs to a joint encoder for the interactions between local features of videos and texts, sacrificing the retrieval efficiency (every text-video pair needs to be fed into the encoder during inference) for the benefits of fine-grained feature learning.

To enable fine-grained video-text interactions and at the same time maintaining high retrieval efficiency, we introduce a novel parametric pretext task for video-text pre-training, namely, \textbf{M}ultiple \textbf{C}hoice \textbf{Q}uestions (\textbf{MCQ}), which properly bridges texts with videos in all their feature levels.
A new module in vitro, termed \textit{BridgeFormer}, makes it possible, as illustrated in Fig.~\ref{fig:framework}.
Based on the backbone of a ``dual-encoder'' framework, BridgeFormer is trained to answer the ``questions'' generated by the text features via visual reasoning with the video features.
MCQ enhances local feature learning within each modality as well as the fine-grained semantic associations cross modalities, and the BridgeFormer can be readily removed when transferring to downstream tasks without the loss of representation discriminativeness.

Specifically, 
we construct the ``questions'' by erasing a content phrase from the raw text, and the correct ``answer'' should be the erased phrase itself.
Motivated by the observation that noun and verb phrases in a text carry rich semantic information~\cite{taco}, 
which can reflect the local objects and object motions in the video respectively,
we randomly choose nouns or verbs as our content phrases.
BridgeFormer is then trained to select the correct answer from multiple choices (all the erased content phrases in a batch) in the form of contrastive learning by resorting to the local features from the video encoder.
Such a proxy training objective enforces the video encoder to capture accurate spatial content (to answer nouns) and temporal dynamics (to answer verbs), promoting the discriminativeness of the local features and the semantic associations between the local video patches and the text phrases.

BridgeFormer connects local features of videos and texts in all feature levels (low-, mid-, and high-level), \textit{i.e.}, taking each stage's features from the video and text encoders as input.
The regularization will be directly imposed on the video and text features, which is different from the video-text feature aggregation by the conventional ``joint-encoder''.
Therefore, the proxy BridgeFormer only serves for the pre-training step and can be seamlessly removed for downstream retrieval, rendering a flexible and efficient model like the conventional ``dual-encoder'' methods, \textit{i.e.}, the similarity between video and text representations can be directly measured via dot product.

\begin{figure}
	\centering
	\includegraphics[width=0.9\linewidth]{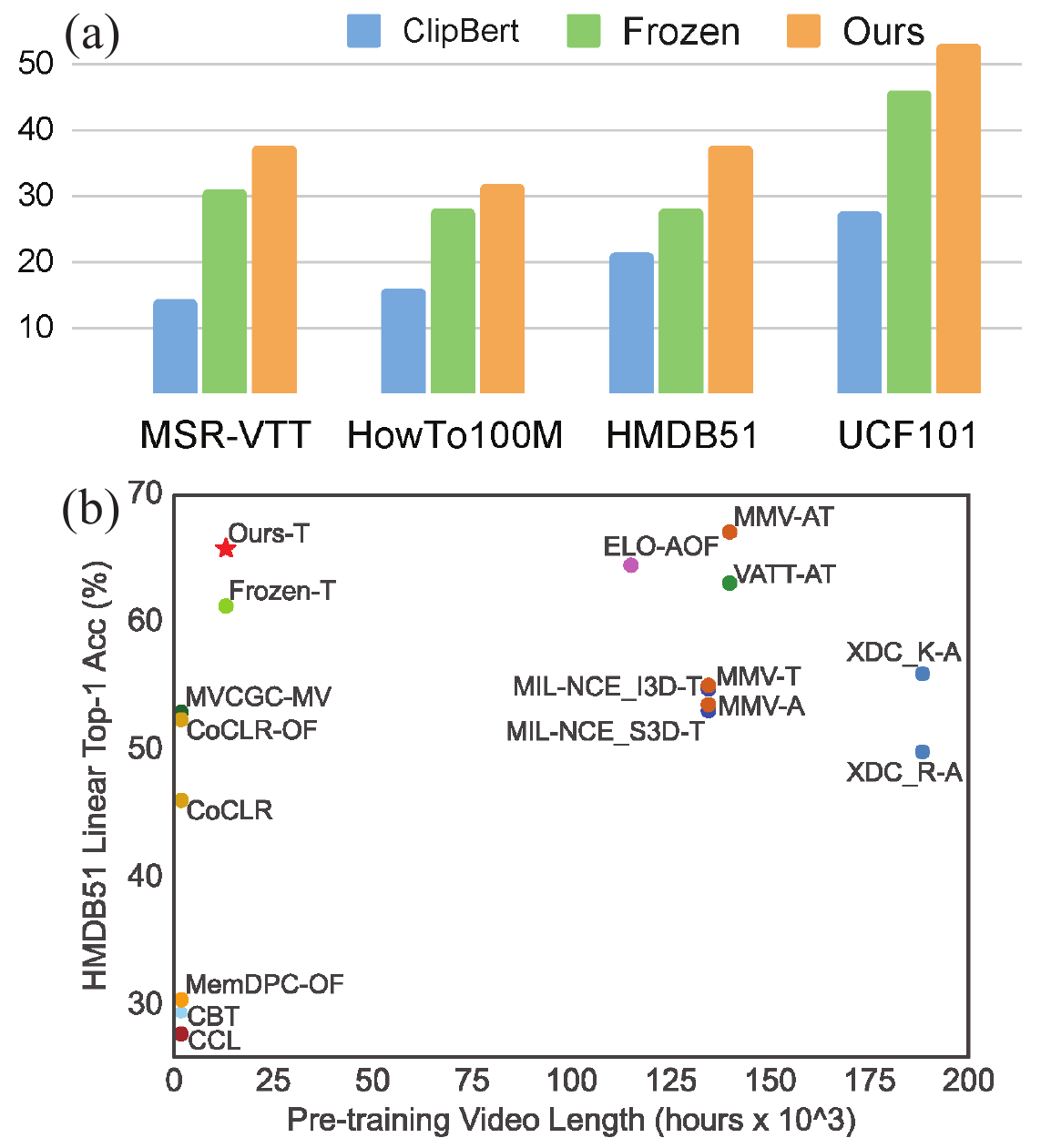}
	\caption{\textbf{(a)} Comparison between recent video-text pre-training methods for zero-shot text-to-video retrieval on MSR-VTT (R@1), HowTo100M (R@50) and zero-shot action recognition (video-to-text retrieval) on HMDB51 (top-1) and UCF101 (top-1). \textbf{(b)} Video length for pre-training and the top-1 accuracy of action recognition with linear evaluation, where ``-X'' denotes the modality used for pre-training besides videos, \textit{i.e.}, optical flow (OF), motion vector (MV), audio (A), and text (T).}
	\vspace{-4mm}	
	\label{fig:hmdb}
\end{figure}

Our contributions are three-fold.
(1) We introduce a novel pretext task, Multiple Choice Questions (MCQ), for video-text pre-training to receive the benefits of both ``dual-encoder'' and ``joint-encoder'' methods, \textit{i.e.}, enhancing fine-grained semantic associations between video and text features at the same time preserving high retrieval efficiency. 
(2) We propose a parametric module, dubbed as BridgeFormer, to realize the pretext task of MCQ, with which the video encoder is trained to be more aware of regional objects and temporal dynamics, and the associations between local video-text features are established.
Since the BridgeFormer will be removed on downstream tasks, 
we do not increase any additional parameters or computational overhead for retrieval compared to vanilla backbones.
(3) Extensive results on text-to-video retrieval with different setups ( \textit{i.e.}, zero-shot and fine-tune) on five datasets, including the large-scale HowTo100M~\cite{howto100m} (1 million videos), demonstrate the large superiority of our method (see Fig.~\ref{fig:hmdb} (a)). 
Furthermore, we evaluate zero-shot action recognition, which can be cast as a video-to-text retrieval task.
Our method significantly surpasses its competitive counterparts by a large margin, as demonstrated in Fig.~\ref{fig:hmdb} (a).
As a bonus, we find our method also benefits single-modality video representations as shown in Fig.~\ref{fig:hmdb} (b), where the top-1 accuracy of action recognition with linear evaluation is reported.
Despite those considerably longer videos being used in state-of-the-art pre-training methods (\textit{e.g.}, 11$\times$ longer in MMV~\cite{MMV} than ours), our method still compares favorably with them.

\section{Related Work}

\paragraph{Pre-training for video-text retrieval.}
Dominant pre-training methods for video-text retrieval can be classified into two categories.
Methods in the first category~\cite{frozen, MIL-NCE, videoclip, taco, coot,avlnet,support, multi,expert} adopt two individual encoders to embed video features and text features, and project them into the same latent space. 
Contrastive objectives~\cite{contrastive1,contrastive2} are used here to distinguish paired video-text data with unpaired data. 
This kind of methods is more favored by large-scale retrieval applications due to its high efficiency.
However, simply 
imposing the regularizations on the final features ([CLS] tokens) from two modalities leads to the insufficient interaction between local video-text representations.
Methods in the second category~\cite{clipbert,videobert,actbert,hero, univl, vlm} ensemble texts and videos as inputs to a joint encoder for the cross-modality fusion, followed by a binary classifier which is trained to predict whether videos and texts are aligned or not. 
Despite they can build local associations between video-text tokens, each pair of video and text candidates needs to be fed into the model for similarity calculation during inference, resulting in extremely low efficiency.
In contrast, our method gains the benefits of the above two kinds of methods, \textit{i.e.}, achieving fine-grained video-text interactions while remaining high retrieval efficiency.

\paragraph{The pretext task of masked word prediction.}
Previous cross-modality pre-training work~\cite{vilt,univl,actbert} use the pretext of masked word prediction (MWP), which randomly masks a proportion of words in the sentence and regularize the network to predict the masked words from a fixed vocabulary under the condition of visual inputs. 
Our introduced MCQ pretext task differs from MWP in two ways: 
(1) Predicting words in MWP imposes the regularizations on low-level word tokens, which may harm the interacted representation learning since the network also needs to serve as a text decoder. In contrast, contrasting answers with content phrases in our MCQ focuses on high-level semantics, showing significantly better results than MWP (will be discussed in experiments).
(2) MCQ erases noun and verb phrases to construct informative questions, which reflects salient semantic information in visual features, while MWP randomly masks words (\eg, function words without content).

\paragraph{Video question answering (VQA).}
Works on video question answering (VQA)~\cite{qa1,qa2,qa3,qa4} aims to answer questions about videos through training a model with question and answer pairs, which cannot be directly applied for pre-training as they are deliberately optimized for increasing VQA accuracy. By contrast, our work aims to learn downstream-agnostic generic features for video-text retrieval, where a new pretext task, multiple choice questions, is proposed to enhance the semantic associations between video and text. Our paper \textit{is the first to} use the form of VQA as a pre-training pretext task, with \textit{two key innovations}: the MCQ loss and the BridgeFormer module. BridgeFormer smoothly bridges the final objective of learning well-aligned video and text features with the regularization of a VQA pretext task.

\paragraph{Video-text retrieval with nouns and verbs.}
Works~\cite{re1,re2,re3,re4} solved video-text retrieval by focusing on verbs and nouns of texts, which are specially designed for retrieval with verbs and nouns as the refined text representations to directly align with videos. By contrast, we  exploit the rich semantics of nouns and verbs in the text to build questions for improving text and video encoders.

\section{Method}
We adopt the ``dual-encoder'' structure for video-text pre-training to realize highly efficient retrieval, 
and propose a new pretext task, Multiple Choice Questions (MCQ), with a parametric module BridgeFormer, to enhance fine-grained semantic associations between videos and texts. 
In this section, we first revisit the dual-encoder in Sec.~\ref{sec:revisit }. We then introduce the pretext task MCQ in Sec.~\ref{sec:mcq} and the pre-training objectives in Sec.~\ref{sec:strategy}. At last, we describe the architecture of three components including a VideoFormer, a TextFormer, and a BridgeFormer in Sec.~\ref{sec:architecture}. 

\subsection{Dual-encoder for Video-text Pre-training: a revisit}\label{sec:revisit }
As shown in Fig.~\ref{fig:method}, we adopt a dual-encoder structure, which consists a VideoFormer for learning video representations from raw video frame pixels, and a TextFormer for encoding text representations from natural languages. Given a video and its corresponding text description (\eg, ``A girl in shorts and a hat is dancing on the green grass''), we first embed their respective representations from VideoFormer and TextFormer, which are projected to a common embedding space as $f_v$ and $f_t$ via two separate linear layers.
The similarity between the video and the text is calculated via the dot product between $f_v$ and $f_t$. A contrastive objective~\cite{contrastive1,contrastive2} is utilized to maximize the similarity between $f_v$ and $f_t$ of positive pairs while minimizing the similarity between $f_v$ and $f_t$ of negative pairs (A video and its corresponding text description is regarded as a positive pair, and otherwise as a negative pair). The independent dual encoder pathways require only the dot product between video and text representations for similarity calculation in retrieval, which ensures the high efficiency.

\subsection{Multiple Choice Questions}\label{sec:mcq}
As shown in Fig.~\ref{fig:framework}, the pretext task MCQ is performed using a parametric module BridgeFormer, which associates all-level intermediate tokens from VideoFormer and TextFormer to answer multiple choice questions. Given observed that noun and verb phrases in a text carry rich semantic information, which can reflect the local objects and object motions in the video respectively, we randomly erase a noun or verb phrase to construct noun or verb questions. BridgeFormer is then trained to select the correct answer from multiple choices (all the erased phrases in a batch) by resorting to the local tokens of VideoFormer in the form of contrastive learning. The pretext task MCQ involves the objectives of answering noun questions and verb questions.

\begin{figure}
	\centering
	\includegraphics[width=1.0\linewidth]{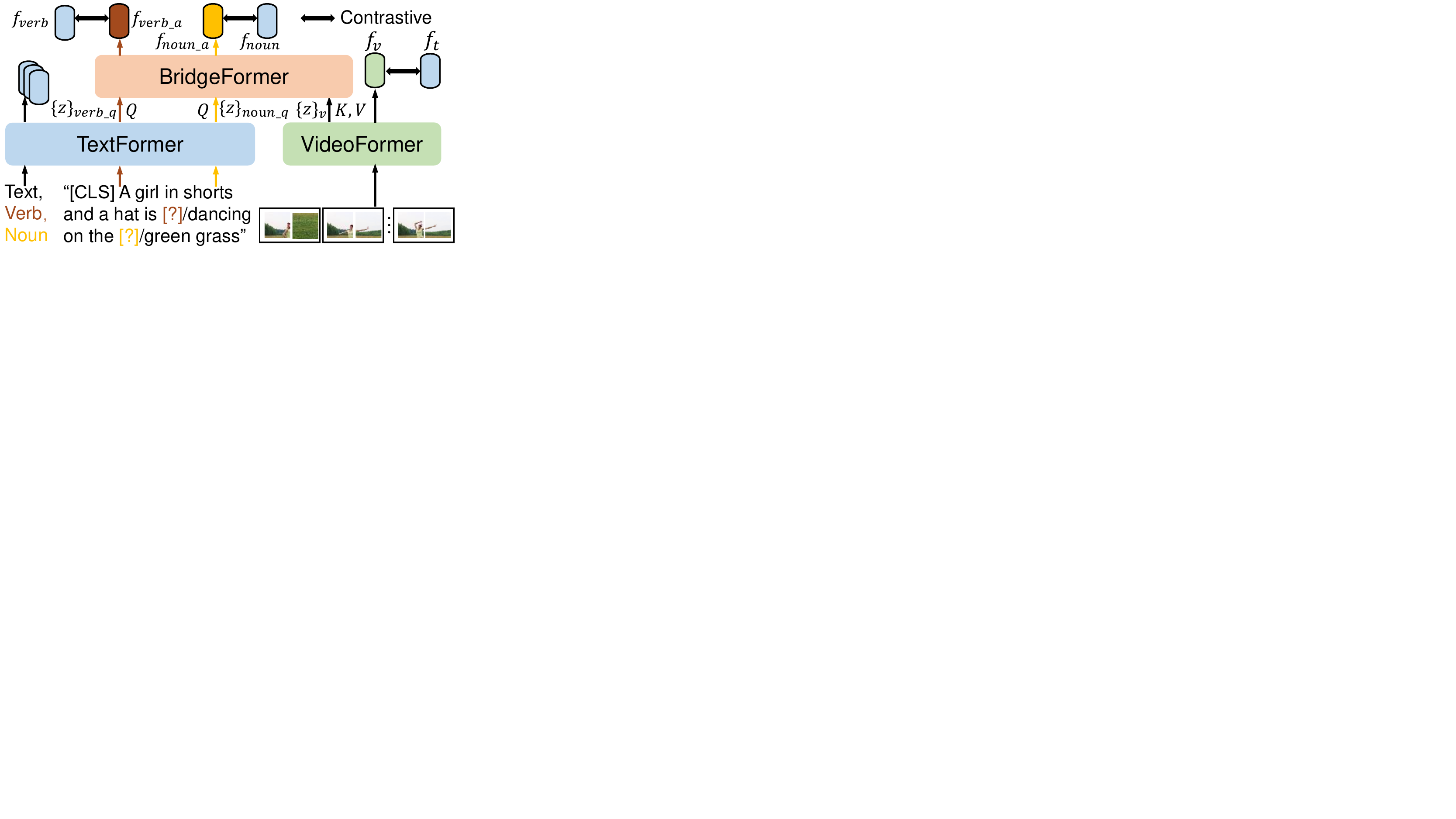}
	\caption{Our pre-training pipeline, which \textbf{(1)} contrasts video representations $f_v$ with text representations $f_t$, \textbf{(2)} trains BridgeFormer to select the correct noun answer by contrasting noun answer representations $f_{noun\_a}$ with noun representations $f_{noun}$, \textbf{(3)} trains BridgeFormer to choose the correct verb answer by contrasting verb answer representations $f_{verb\_a}$ with verb representations $f_{verb}$. Note that BridgeFormer receives all-level tokens as the input, but we only draw one pathway here for brevity.}
	\vspace{-4mm}	
	\label{fig:method}
\end{figure}

{\flushleft \bf Answer Noun Question.} Given a video and its corresponding text description (\eg, ``A girl in shorts and a hat is dancing on the green grass''), we randomly erase a noun phrase (\eg, ``green grass'') as a noun question (\eg, ``A girl in shorts and a hat is dancing on the [?]''). As shown in Fig.~\ref{fig:method}, the noun question is fed into TextFormer for intermediate text tokens $\{z\}_{noun\_q}$. The intermediate video tokens are extracted from VideoFormer as $\{z\}_v$.  BridgeFormer takes the noun question tokens $\{z\}_{noun\_q}$ as the query, and video tokens $\{z\}_v$ as the key and value to obtain the noun answer representations through cross-modality attention. The erased noun phrase is fed into TextFormer for noun representations.  Similarly, the noun answer representations and the noun representations are projected into a common embedding space as $f_{noun\_a}$ and $f_{noun}$ via two separate linear layers, and their similarity is calculated via dot product. We adopt a contrastive objective to maximize the similarity between $f_{noun\_a}$ and $f_{noun}$, when $f_{noun}$ is the representations of the correct noun phrase, and minimize the similarity between  $f_{noun\_a}$ and $f_{noun}$, when $f_{noun}$ is the representations of other (wrong) noun phrases.  Training BridgeFormer to select the correct noun phrase by resorting to video tokens enforces VideoFormer to capture accurate spatial content.

{\flushleft \bf Answer Verb Question.} Similarly,  we randomly erase a verb phrase (\eg, ``dancing'') of the text description as a verb question (\eg, ``A girl in shorts and a hat is [?] on the green grass''). As shown in Fig.~\ref{fig:method}, BridgeFormer takes verb question text tokens $\{z\}_{verb\_q}$ from TextFormer as the query, and video tokens $\{z\}_v$ as the key and value to obtain the verb answer representations. The erased verb phrase is fed into TextFormer for verb representations. The verb answer representations and the verb representations are projected into a common embedding space as $f_{verb\_a}$ and $f_{verb}$. A contrastive objective is adopted to maximize the similarity between $f_{verb\_a}$ and $f_{verb}$, when $f_{verb}$ is the representations of the correct verb phrase, and minimize the similarity between $f_{verb\_a}$ and $f_{verb}$, when $f_{verb}$ is the representations of other verb phrases. Training BridgeFormer to choose the correct verb phrase through seeking help from video tokens forces VideoFormer to capture detailed temporal dynamics.

\subsection{Pre-training Objectives}\label{sec:strategy}
 We adopt the Noise-Contrastive Estimation (NCE)~\cite{contrastive1,contrastive2} as the contrastive objective and combine three objectives to optimize the entire model in an end-to-end manner as follows,
\begin{small}
\begin{equation}
	\mathcal{L} = \mathcal{L}_\text{vanilla} + \mathcal{L}_\text{noun} + \mathcal{L}_\text{verb}
\end{equation}
\end{small}
where $\mathcal{L}_\text{vanilla}$ is the NCE loss between video representations $f_v$ and text representations $f_t$, $\mathcal{L}_\text{noun}$ is the NCE loss between noun answer representations $f_{noun\_a}$ and noun representations $f_{noun}$, $\mathcal{L}_\text{verb}$ is the NCE loss between verb answer representations $f_{verb\_a}$ and verb representations $f_{verb}$. We formulate NCE loss as below,
	\begin{small}
		\begin{equation}
			\textbf{NCE}(x_i,y_i)= - \text{log}\frac{\text{exp}(x_i^Ty_i/\tau)}{\sum_{j=1}^B\text{exp}(x_i^Ty_j/\tau)} 
		\end{equation}
	\end{small}
where $B$ is the number of the batch size and the temperature hyper-parameter $\tau$ is empirically set to $0.05$ per \cite{frozen}.

\subsection{Model Architecture}\label{sec:architecture}
\subsubsection{VideoFormer}
{\flushleft \bf Input.} VideoFormer takes a video $V\in R^{M \times 3 \times H \times W}$ as input containing variable $M$ frames of resolution $H \times W$. The input video is first divided into $M \times N$ patches of size $P \times P$, where $N=HW/P^2$. The video patches $v\in R^{M \times 3 \times N \times P \times P}$ are fed into a linear projection head with a convolutional layer and are flattened into a sequence of tokens $z_v \in R^{M\times N \times D}$, where $D$ is the number of embedding dimensions. Following BERT~\cite{bert}, a learnable [CLS] token is concatenated to the beginning of the token sequence, which is used to produce the final video representations. Learnable spatial positional embeddings $E_{pos} \in R^{(N+1)\times D}$ are added to each video token as the final input token sequence  $z_v^0 \in R^{(1+M\times N) \times D}$ and all patches in the same spatial location in different frames are given the same spatial positional embedding.

{\flushleft \bf VideoBlock.} The input video token sequence $\{z\}_v^0$ is fed into VideoFormer, which consists of a stack of VideoBlocks as shown in Fig.~\ref{fig:architecture}, adopting the structure of ViT~\cite{vit}. We make a minor modification to the original ViT to allow for the input of video frames with variable length. Specifically, given $z_v^{l-1} \in R^{(1+M\times N) \times D}$ from previous VideoBlock, we perform multi-head attention (MSA)~\cite{vit} for the [CLS] token through attending to all $(1+M\times N)$ patches across time and space for temporal and spatial self-attention. For the rest $(M\times N)$ patch tokens, MSA is performed within each of $M$ frames with $N+1$ tokens ($N$ patch tokens and 1 [CLS] token) for spatial self-attention. The video representations are obtained from the [CLS] token of the final VideoBlock.

\subsubsection{TextFormer}
{\flushleft \bf Input.} TextFormer takes three kinds of nature languages as inputs, including a complete text description, noun or verb questions with a noun or verb phrase erased, and the erased noun or verb phrase. A [CLS] token is concatenated to the beginning of the input for final text representations.

\begin{figure}
	\centering
	\includegraphics[width=0.9\linewidth]{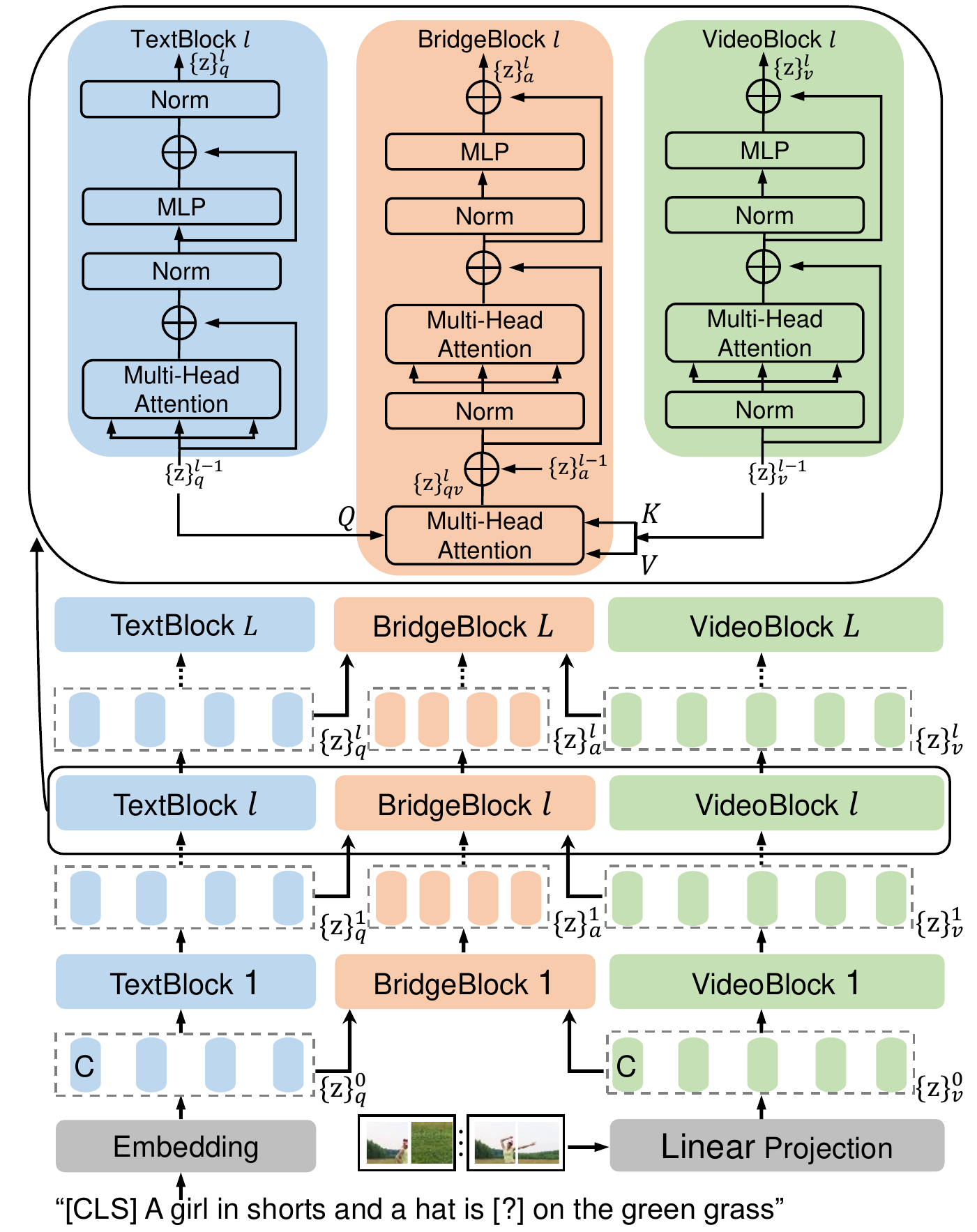}
	\caption{The architecture of TextFormer, VideoFormer and BridgeFormer, which contain a stack of TextBlocks,
	VideoBlocks and BridgeBlocks respectively. Tokens from all-level VideoBlock and TextBlock are fed into the corresponding
	BridgeBlock to perform cross-modal attention and then are added to the output tokens of the previous BridgeBlock (if any). Each block performs a series of operations such as multi-head attention~\cite{vit}, normalization (norm) and multi-layer perception~\cite{bert} (MLP).}
	\vspace{-4mm}	
	\label{fig:architecture}
\end{figure}

{\flushleft \bf TextBlock.} We adopt a multi-layer bidirectional transformer encoder~\cite{distilbert} as TextFormer, which consists of a stack of TextBlocks as shown in Fig.~\ref{fig:architecture}. 

\subsubsection{BridgeFormer}
{\flushleft \bf Input.} BridgeFormer takes noun question or verb question tokens from TextFormer as the query, and video tokens from VideoFormer as the key and value to obtain the answer representations with cross-modality attention.

{\flushleft \bf BridgeBlock.} BridgeFormer is built upon a vision transformer with a stack of BridgeBlocks as shown in Fig.~\ref{fig:architecture}. Specifically, given noun question or verb question text tokens $\{z\}_q^{l-1}\in R^{L\times D}$ from TextBlock as the query, and video tokens $\{z\}_v^{l-1}\in R^{M\times (N\times D)}$ (without the [CLS] token) from VideoBlock as the key and value, BridgeBlock-$l$ obtains the interacted tokens $\{z\}_{qv}^l$ through performing multi-head attention, which calculates the cross-modality attention between the question text tokens and video patch tokens within each frame. The interacted tokens $\{z\}_{qv}^l$ added with the output $\{z\}_a^{l-1}$ from the previous BridgeBlock further go through the attention block for temporal and spatial self-attention as shown in Fig.~\ref{fig:architecture} to obtain the answer tokens $\{z\}_a^l$. The answer representations are extracted from the [CLS] token of the final block.

\begin{table*}\centering
	\vspace{-10pt}
	\caption{Experiments of text-to-video retrieval on MSR-VTT test set with 1K videos, where \textbf{higher} R@k and \textbf{lower} MedR (Median Rank) indicate better performance. \textbf{Video Encoder Input}: 3D features from the architectures (Raw Videos means training on raw video frame pixels without using pre-extracted features). \textbf{$\#$ Pairs PT}: the number of video-text pairs for pre-training. We show results with zero-shot evaluation (top) and fine-tuning evaluation (bottom).} 
	\vspace{-5pt} 
	\scalebox{0.8}{
		\begin{tabular}{c|cccccccc}
			\toprule[1pt]
			\multirow{1}*{Method}&Year&Video Encoder Input&PT Dataset&$\#$Pairs PT&R@1&R@5&R@10&MedR\\
			\hline
			\multirow{1}*{ActBERT~\cite{actbert}}&2020&ResNet-3D&HowTo100M&120M&8.6&23.4&33.1&36.0\\
			\multirow{1}*{MMV~\cite{MMV}}&2020&Raw Videos&HowTo100M, AudioSet&138M&9.3&23.0&31.1&38.0\\
			\multirow{1}*{MIL-NCE~\cite{MIL-NCE}}&2020&Raw Videos&HowTo100M&120M&9.9&24.0&32.4&29.6\\
			\multirow{1}*{VATT~\cite{vatt}}&2021&Raw Videos&HowTo100M, AudioSet&138M&-&-&29.7&49.0\\
			\multirow{1}*{NoiseEst~\cite{noise}}&2021&ResNeXt-101&HowTo100M&110M&8.0&21.3&29.3&33.0\\
			\multirow{1}*{TACo~\cite{taco}}&2021&I3D, S3D&HowTo100M&120M&9.8&25.0&33.4&29.0\\
			\multirow{1}*{VideoCLIP~\cite{videoclip}}&2021&S3D&HowTo100M&110M&10.4&22.2&30.0&-\\
			\multirow{1}*{MCN~\cite{mcn}}&2021&ResNeXt-101&HowTo100M&120M&10.5&25.2&33.8&-\\	
			\multirow{1}*{SupportSet~\cite{support}}&2021&R(2+1)D-34&HowTo100M&120M&12.7&27.5&36.2&24.0\\
			\multirow{1}*{Frozen~\cite{frozen}}&2021&Raw Videos&CC3M, WebVid-2M&5.5M&18.7&39.5&51.6&10.0\\
			\multirow{1}*{AVLnet~\cite{avlnet}}&2021&ResNeXt-101&HowTo100M&120M&19.6&40.8&50.7&9.0\\
			\multirow{1}*{Ours}&2021&Raw Videos&CC3M, WebVid-2M&5.5M&\textbf{26.0}&\textbf{46.4}&\textbf{56.4}&\textbf{7.0}\\
			\toprule[1pt]	
			\multirow{1}*{ActBERT~\cite{actbert}}&2020&ResNet-3D&HowTo100M&120M&16.3&42.8&56.9&10.0\\
			\multirow{1}*{UniVL~\cite{univl}}&2020&S3D&HowTo100M&110M&21.2&49.6&63.1&6.0\\
			\multirow{1}*{MMT~\cite{multi}}&2020&S3D&HowTo100M&120M&26.6&57.1&69.6&4.0\\
			\multirow{1}*{HERO~\cite{hero}}&2021&SlowFast&TV and HowTo100M&120M&16.8&43.4&57.7&-\\	
			\multirow{1}*{NoiseEst~\cite{noise}}&2021&ResNeXt-101&HowTo100M&110M&17.4&41.6&53.6&8.0\\
			\multirow{1}*{ClipBert~\cite{clipbert}}&2021&Raw Videos&COCO, VisGenome&5.6M&22.0&46.8&59.9&6.0\\
			\multirow{1}*{AVLnet~\cite{avlnet}}&2021&ResNeXt-101&HowTo100M&120M&27.1&55.6&66.6&4.0\\
			\multirow{1}*{VLM~\cite{vlm}}&2021&S3D&HowTo100M&110M&28.1&55.5&67.4&4.0\\
			\multirow{1}*{TACo~\cite{taco}}&2021&I3D, S3D&HowTo100M&120M&28.4&57.8&71.2&4.0\\
			\multirow{1}*{SupportSet~\cite{support}}&2021&R(2+1)D-34&HowTo100M&120M&30.1&58.5&69.3&3.0\\	
			\multirow{1}*{VideoCLIP~\cite{videoclip}}&2021&S3D&HowTo100M&110M&30.9&55.4&66.8&-\\
			\multirow{1}*{Frozen~\cite{frozen}}&2021&Raw Videos&CC3M, WebVid-2M&5.5M&31.0&59.5&70.5&3.0\\
			\multirow{1}*{Ours}&2021&Raw Videos&CC3M,  WebVid-2M&5.5M&\textbf{37.6}&\textbf{64.8}&\textbf{75.1}&\textbf{3.0}\\			
			\bottomrule[1pt]
	\end{tabular}}
	\vspace{-5pt}
	\label{tab:msrvtt}
\end{table*}

\section{Experiments}
\subsection{Pre-training Datasets}
Following the recent work~\cite{frozen}, we jointly pre-train our model on an image dataset Google Conceptual Captions (CC3M)~\cite{cc3m} with 3.3M image-text pairs, and a video dataset WebVid-2M~\cite{frozen} with 2.5M video-text pairs. We do not pre-train our model on the large-scale video-text dataset HowTo100M~\cite{howto100m} with 136M video-text pairs considering the enormous computation cost. Instead, we use HowTo100M as a large-scale zero-shot text-to-video retrieval benchmark for evaluation, which is in line with real-world applications.

\subsection{Downstream Tasks}
{\flushleft \bf Text-to-Video Retrieval.} (a). \textbf{MSR-VTT}~\cite{msr} contains 10K YouTube videos with 200K
descriptions, which is split into 9K videos for training and 1K videos for test.
(b). \textbf{MSVD}~\cite{msvd} consists of 1,970 videos from YouTube with 80K descriptions, which is split into 1200, 100 and 670 videos for training, validation and testing.
(c). \textbf{LSMDC}~\cite{lsmdc} consists of 118,081 video clips from 202 movies. The validation set and the test set contain 7,408 and 1,000 videos. 
(d). \textbf{DiDeMo}~\cite{didemo} contains 10K Flickr videos with 40K sentences, where the test set contains 1,000 videos. We concatenate all sentence descriptions for a video as a single query following~\cite{frozen}.
(e). \textbf{HowTo100M}~\cite{howto100m} contains 1.22M videos with 136M descriptions. All sentence descriptions for a video are concatenated as a single query. To our knowledge, it is the first time that downstream text-to-video retrieval is evaluated on the large-scale dataset, \ie, HowTo100M. 
Two setting are explored for evaluation, including \textbf{zero-shot} and \textbf{fine-tune}.

{\flushleft \bf Action Recognition.}
(a). \textbf{HMDB51}~\cite{hmdb}, which contains 6,766 videos with 51 categories.
(b). \textbf{UCF101}~\cite{ucf}, which contains 13,320 videos with 101 action classes. 
Three setting are explored for evaluation, including \textbf{linear}, where parameters of the learned video encoder are frozen and only a linear classifier is optimized, \textbf{fine-tune}, where the video encoder is fine-tuned with the linear classifier, and \textbf{zero-shot}, which performs video-to-text retrieval through using the names of the action classes as the text description.

\begin{table*}[t]
	\vspace{1em}
	\centering
	\caption{Experiments of text-to-video retrieval on different datasets, where \textbf{higher} R@k and \textbf{lower} MedR (Median Rank) indicate better performance. We show results with zero-shot evaluation (top) and fine-tuning evaluation (bottom).}
	\vspace{-0.5em}
	\subfloat[
	MSVD test set with 670 videos.
	\label{tab:msvd}
	]{
		\centering
		\begin{minipage}{0.29\linewidth}{\begin{center}
					\tablestyle{1pt}{1.05}
					\begin{tabular}{c|cccc}
						\toprule[1pt]
						\multirow{1}*{Method}&R@1&R@5&R@10&MedR\\
						\hline
						\multirow{1}*{NoiseEst~\cite{noise}}&13.7&35.7&47.7&12.0\\
						\multirow{1}*{SupportSet~\cite{support}}&21.4&46.2&57.7&6.0\\
						\multirow{1}*{Frozen~\cite{frozen}}&33.7&64.7&76.3&3.0\\
						\multirow{1}*{Ours}&\textbf{43.6}&\textbf{74.9}&\textbf{84.9}&\textbf{2.0}\\
						\toprule[1pt]	
						\multirow{1}*{NoiseEst~\cite{noise}}&20.3&49.0&63.3&6.0\\
						\multirow{1}*{SupportSet~\cite{support}}&28.4&60.0&72.9&4.0\\
						\multirow{1}*{Frozen~\cite{frozen}}&45.6&79.8&88.2&2.0\\
						\multirow{1}*{Ours}&\textbf{52.0}&\textbf{82.8}&\textbf{90.0}&\textbf{1.0}\\	
						\bottomrule[1pt]
					\end{tabular}
		\end{center}}\end{minipage}
	}
	\hspace{2em}
	\subfloat[
	LSMDC test set with 1K videos.
	\label{tab:lsmdc}
	]{
		\begin{minipage}{0.29\linewidth}{\begin{center}
					\tablestyle{1pt}{1.05}
					\begin{tabular}{c|cccc}
						\toprule[1pt]
						\multirow{1}*{Method}&R@1&R@5&R@10&MedR\\
						\hline
						\multirow{1}*{AVLnet~\cite{avlnet}}&1.4&5.9&9.4&273.5\\
						\multirow{1}*{NoiseEst~\cite{noise}}&4.2&11.6&17.1&119.0\\
						\multirow{1}*{Frozen~\cite{frozen}}&9.3&22.0&30.1&51.0\\
						\multirow{1}*{Ours}&\textbf{12.2}&\textbf{25.9}&\textbf{32.2}&\textbf{42.0}\\	
						\toprule[1pt]	
						\multirow{1}*{NoiseEst~\cite{noise}}&6.4&19.8&28.4&39.0\\
						\multirow{1}*{MMT~\cite{multi}}&12.9&29.9&40.1&19.3\\
						\multirow{1}*{Frozen~\cite{frozen}}&15.0&30.8&39.8&20.0\\
						\multirow{1}*{Ours}&\textbf{17.9}&\textbf{35.4}&\textbf{44.5}&\textbf{15.0}\\
						\bottomrule[1pt]
					\end{tabular}
		\end{center}}\end{minipage}
	}
	\hspace{2em}
	\subfloat[
	DiDeMo test set with 1K videos.
	\label{tab:didemo}
	]{
		\begin{minipage}{0.29\linewidth}{\begin{center}
					\tablestyle{1pt}{1.05}
					\begin{tabular}{c|cccc}
						\toprule[1pt]
						\multirow{1}*{Method}&R@1&R@5&R@10&MedR\\
						\hline
						\multirow{1}*{VideoCLIP~\cite{videoclip}}&16.6&46.9&-&-\\
						\multirow{1}*{Frozen~\cite{frozen}}&21.1&46.0&56.2&7.0\\
						\multirow{1}*{Ours}&\textbf{25.6}&\textbf{50.6}&\textbf{61.1}&\textbf{5.0}\\
						\toprule[1pt]		
						\multirow{1}*{HERO~\cite{hero}}&2.1&-&11.4&-\\
						\multirow{1}*{CE~\cite{expert}}&16.1&41.1&82.7&8.3\\
						\multirow{1}*{ClipBert~\cite{clipbert}}&20.4&48.0&60.8&6.0\\
						\multirow{1}*{Frozen~\cite{frozen}}&31.0&59.8&72.4&3.0\\
						\multirow{1}*{Ours}&\textbf{37.0}&\textbf{62.2}&\textbf{73.9}&\textbf{3.0}\\
						\bottomrule[1pt]
					\end{tabular}
		\end{center}}\end{minipage}
	}
	\label{tab:others} \vspace{-.5em}
\end{table*}

\begin{table}\centering
	\caption{Experiments of zero-shot text-to-video retrieval on the large-scale HowTo100M, where \textbf{higher} R@k and \textbf{lower} MedR indicate better performance. ``Video Num'' denotes the number of sampled videos for evaluation, where 1M denotes the whole set.} 
	\vspace{-5pt} 
	\scalebox{0.8}{
		\begin{tabular}{c|ccccc}
			\toprule[1pt]
			\multirow{1}*{Video Num}&Method&R@50&R@200&R@500&MedR\\
			\hline
			\multirow{3}*{10K}&ClipBert~\cite{clipbert}&15.8&33.6&49.8&506.0\\
			&Frozen~\cite{frozen}&28.0&46.6&61.5&244.0\\
			&Ours&\textbf{31.6}&\textbf{50.9}&\textbf{65.2}&\textbf{189.0}\\	
			\hline
			\multirow{2}*{50K}&Frozen~\cite{frozen}&13.4&25.0&36.2&1247.0\\
			&Ours&\textbf{15.9}&\textbf{28.6}&\textbf{40.2}&\textbf{965.0}\\	
			\hline
			\multirow{2}*{0.1M}&Frozen~\cite{frozen}&9.4&18.5&27.5&2519.0\\
			&Ours&\textbf{11.5}&\textbf{21.7}&\textbf{31.2}&\textbf{1907.0}\\	
			\hline
			\multirow{2}*{0.5M}&Frozen~\cite{frozen}&4.0&8.5&13.4&12501.0\\
			&Ours&\textbf{5.0}&\textbf{10.3}&\textbf{15.9}&\textbf{9449.0}\\	
			\hline
			\multirow{2}*{1M}&Frozen~\cite{frozen}&2.6&5.9&9.5&24597.0\\
			&Ours&\textbf{3.4}&\textbf{7.3}&\textbf{11.6}&\textbf{18612.0}\\	
			\bottomrule[1pt]
	\end{tabular}}
	\vspace{-10pt}
	\label{tab:howto}
\end{table}

\subsection{Implementation Details}
Videos are resized to 224 $\times$ 224 as input. We divide a video into $M$ equal segments, and randomly sample a single frame from each segment for training while uniformly sample a frame from each segment for testing. VideoFormer contains 12 blocks with patch size $P=16$, and sequence dimension $D=768$. It is initialized with ViT~\cite{vit} weights trained on ImageNet-21k following~\cite{frozen}. TextFormer adopts the architecture of DistilBERT~\cite{distilbert} pre-trained on English Wikipedia and Toronto Book Corpus.  The dimension of the common feature space is set to 256. The temperature hyper-parameter of the contrastive objective is set to 0.05. The above implementation details follow the recent work ~\cite{frozen} for fair comparison. BridgeFormer contains 12 blocks. We first pre-train our model on the image dataset CC3M and video dataset WebVid-2M using 1 frame for 10 epochs with the batch size of 2048 and the learning rate of $1\times {10}^{-4}$. We then pre-train our model on the video dataset WebVid-2M using 4 frames for 4 epochs with the batch size of 800 and the learning rate of $3\times {10}^{-5}$. Pre-training takes a total of 25 hours. For downstream tasks, 4 frames for text-to-video retrieval and 16 frames for action recognition are uniformly sampled following the setting of previous work~\cite{frozen, MIL-NCE}.

\subsection{Main Results}
\subsubsection{Text-to-Video Retrieval}
Table.~\ref{tab:msrvtt} lists the results on MST-VTT~\cite{msr}. First of all, our method outperforms all previous work by a large margin. The significantly higher performance of our model under the zero-shot evaluation demonstrates the stronger generalization ability of our pre-trained model. Fine-tuning our pre-trained model on the training set of MSR-VTT also surpasses its counterparts overwhelmingly, showing its advantage in using task-specific data for optimization. 
Second, while previous work mostly pre-train on HowTo100M~\cite{howto100m} with the magnitude exceedingly large than our pre-training dataset CC3M~\cite{cc3m} and WebVid-2M~\cite{frozen} (20x larger in the number of video-text pairs), our method still achieves the highest performance with much lower computation cost (\ie VATT~\cite{vatt} takes 3 days using 256 TPUs while ours takes 25 hours using 40 A100.)
Third, previous work rely on pre-extracted features from ``expert'' models as the input of the video encoder (\ie SupportSet~\cite{support} uses features from a 34-layer,
R(2+1)-D model~\cite{3d} pre-trained on IG65M~\cite{ig65} as the input), while our model takes raw video frame pixels as inputs and achieves significant performance gain.
Finally, compared with previous work~\cite{clipbert, hero, univl, vlm, actbert} that adopt a joint encoder to concatenate videos and texts as inputs and thus every text-video combination needs to be imputed to the model for retrieval, our model only contains a video and a text encoder for downstream retrieval, which requires only the dot product between the video and text representations, thus greatly improves efficiency. 
We further show text-to-video retrieval results on MSVD~\cite{msvd}, DiDeMo~\cite{didemo} and LSMDC in Table.~\ref{tab:others}. We can observe that our model achieves the best performance on these three datasets with both zero-shot and fine-tuning evaluation.

Besides evaluating text-to-video retrieval on a relatively small number of videos following previous work (\textit{e.g.} 1K videos in MSR-VTT test set), 
we evaluate our model on the large-scale HowTo100M with 1 million videos, which is a more challenging and realistic scenario. Table.~\ref{tab:howto} shows that our pre-trained model surpasses SOTA Frozen~\cite{frozen}, ranging from 10K videos to 1M videos. Since our method and Frozen both adopt two encoders (built on ViT~\cite{vit} and DistilBERT~\cite{distilbert}) for retrieval and are pre-trained on the same datasets, the superior performance of ours proves the effectiveness of our pretext task MCQ in learning powerful representations for text-to-video retrieval.

\begin{table}\centering
	\vspace{8pt}
	\caption{Experiments of zero-shot action recognition (video-to-text retrieval) on HMDB51 and UCF101, in terms of top-1 accuracy. ``S'' denotes different test splits and ``Mean'' reports the results averaged on three splits.} 
	\vspace{-5pt}
	\scalebox{0.75}{
		\begin{tabular}{c|cccc|cccc}
			\toprule[1pt]
			Method&\multicolumn{4}{c|}{HMDB51}&\multicolumn{4}{c}{UCF101}\\
			&S1&S2&S3&Mean&S1&S2&S3&Mean\\
			\hline
			ClipBert~\cite{clipbert}&20.0&22.0&22.3&21.4&27.5&27.0&28.8&27.8\\
			Frozen~\cite{frozen}&27.5&28.3&27.7&27.8&45.4&44.7&47.7&45.9\\
			Ours&\textbf{38.0}&\textbf{36.1}&\textbf{39.1}&\textbf{37.7}&\textbf{51.1}&\textbf{54.3}&\textbf{53.8}&\textbf{53.1}\\	
			\bottomrule[1pt]
	\end{tabular}}
	\vspace{-10pt}
	\label{tab:zero_action}
\end{table}

\subsubsection{Action Recognition}
We conduct zero-shot action recognition on HMDB51~\cite{hmdb} and UCF101~\cite{ucf}, which can be treated as \textbf{video-to-text retrieval} and it is not evaluated in recent methods. As shown in Table.~\ref{tab:zero_action}, our model significantly surpasses its competitive counterparts. The top-1 accuracy of our model averaged on three splits improves 16.3\% and 9.9\% on HMDB51, 25.3\% and 7.2\% on UCF101 than the recently proposed ClipBert and Frozen, which shows the great advantage of our model in learning joint representations between videos and languages that enable zero-shot action recognition.

We further evaluate the \textbf{single-modality video representations} of our model via action recognition with linear and fully fine-tuning evaluation as shown in Table.~\ref{tab:action}, where the representations from VideoFormer are extracted as the input of a trainable linear classifier. 
Our method achieves higher accuracy than some previous work that pre-train their model on datasets with considerably longer video time (\textit{e.g.} 14$\times$ longer in XDC~\cite{XDC}, 10$\times$ longer in MIL-NCE~\cite{MIL-NCE} and VATT~\cite{vatt}), showing the effectiveness of our method in learning transferable video representations for action recognition.
Despite MMV~\cite{MMV} performs better than our method when pre-training on datasets 11$\times$ longer than ours with multiple modalities including audio and text besides video, its performance lags far behind ours when only audio and video or text and video are used. We can conclude that our method utilizes the language modality more efficiently to learn stronger video representations with fewer video hours.

\begin{table}\centering
	\caption{Experiments of action recognition on HMDB51 and UCF101 with linear evaluation (Lin) and fully fine-tuning evaluation (Full). The evaluation metric is top-1 accuracy. ``Mod'' denotes the modality used for pre-training besides videos, \textit{i.e.}, optical flow (OF), motion vector (MV), audio (A), text (T). ``Len'' denotes the video length for pre-training in \textit{kilo} hours.} 
	\vspace{-5pt}
	\scalebox{0.8}{
		\begin{tabular}{c|cc|cc|cc}
			\toprule[1pt]
			\multirow{1}*{Method}&Mod&Len (K)&\multicolumn{2}{c|}{HMDB}&\multicolumn{2}{c}{UCF}\\
			&&&Lin&Full&Lin&Full\\
			\hline
			\multirow{1}*{CCL~\cite{CCL}}&-&1.8&29.5&37.8&54.0&69.4\\
			\multirow{1}*{CBT~\cite{CBT}}&-&1.8&29.5&44.5&54.0&79.5\\
			\multirow{1}*{MemDPC~\cite{MemDPC}}&OF&1.8&30.5&54.5&54.1&86.1\\
			\multirow{1}*{CoCLR~\cite{COCLR}}&OF&1.8&52.4&62.9&77.8&90.6\\
			\multirow{1}*{MVCGC}&MV&1.8&53.0&63.4&78.0&90.8\\
			\multirow{1}*{XDC$\_$R~\cite{XDC}}&A&188.3&49.9&61.2&80.7&88.8\\
			\multirow{1}*{XDC$\_$K~\cite{XDC}}&A&188.3&56.0&63.1&85.3&91.5\\
			\multirow{1}*{MIL-NCE~\cite{MIL-NCE}}&T&134.5&54.8&59.2&83.4&89.1\\
			\multirow{1}*{Frozen~\cite{frozen}}&T&13.0&61.3&66.3&87.8&89.8\\	
			\multirow{1}*{VATT~\cite{vatt}}&A, T&139.8&63.3&-&89.2&-\\
			\multirow{1}*{ELO~\cite{elo}}&A, OF&115.0&64.5&67.4&-&93.8\\
			\multirow{1}*{MMV~\cite{MMV}}&A&134.5&53.6&-&77.1&-\\
			\multirow{1}*{MMV~\cite{MMV}}&T&134.5&55.1&-&86.8&-\\
			\multirow{1}*{MMV~\cite{MMV}}&A, T&139.8&\textbf{67.1}&\textbf{75.0}&\textbf{91.8}&\textbf{95.2}\\
			\multirow{1}*{Ours}&T&13.0&65.8&69.8&89.1&92.3\\	
			\bottomrule[1pt]
	\end{tabular}}
	\vspace{-10pt}
	\label{tab:action}
\end{table}

\begin{table*}\centering\small
	\caption{Text-to-video retrieval results of models initialized from CLIP~\cite{clip} weights on different datasets under zero-shot and fine-tune evaluation, where \textbf{higher} R@k and \textbf{lower} MdR (Median Rank) and MnR (Mean Rank) indicate better performance.} 
	\vspace{-5pt} 
	\scalebox{0.8}{
		\begin{tabular}{c|ccccc|ccccc|ccccc}
			\toprule[1pt]
			&\multicolumn{5}{c|}{MSR-VTT}&\multicolumn{5}{c|}{MSVD}&\multicolumn{5}{c}{LSMDC}\\
			\multirow{1}*{Method}&R@1&R@5&R@10&MdR&MnR&R@1&R@5&R@10&MdR&MnR&R@1&R@5&R@10&MdR&MnR\\
			\hline
			CLIP-straight~\cite{straight}&31.2&53.7&64.2&4.0&-&37.0&64.1&73.8&3.0&-&11.3&22.7&29.2&56.5&-\\
			CLIP4Clip~\cite{clip4clip}&32.0&57.0&66.9&4.0&34.0&38.5&66.9&76.8&2.0&17.8&15.1&28.5&36.4&28.0&117.0\\
			Ours&\textbf{33.2}&\textbf{58.0}&\textbf{68.6}&\textbf{4.0}&\textbf{25.7}&\textbf{48.4}&\textbf{76.4}&\textbf{85.8}&\textbf{2.0}&\textbf{7.4}&\textbf{15.5}&\textbf{30.7}&\textbf{38.7}&\textbf{22.0}&\textbf{97.9}\\
			\bottomrule[1pt]
			CLIP4Clip~\cite{clip4clip}&43.1&70.4&\textbf{80.8}&2.0&16.2&46.2&76.1&84.6&2.0&10.0&20.7&38.9&47.2&13.0&65.3\\
			Ours&\textbf{44.9}&\textbf{71.9}&80.3&\textbf{2.0}&\textbf{15.3}&\textbf{54.4}&\textbf{82.8}&\textbf{89.4}&\textbf{1.0}&\textbf{6.1}&\textbf{21.8}&\textbf{41.1}&\textbf{50.6}&\textbf{10.0}&\textbf{60.5}\\
			\bottomrule[1pt]
	\end{tabular}}
	\vspace{-5pt}
	\label{tab:clip}
\end{table*}

\begin{table}\centering\small
	\caption{Ablation studies on different components of MCQ. Results of zero-shot text-to-video retrieval on MSR-VTT and zero-shot action recognition on HMDB51 and UCF101 are reported.} 
	\vspace{-5pt} 
	\scalebox{0.8}{
		\begin{tabular}{c|ccc|cc}
			\toprule[1pt]
			&\multicolumn{3}{c|}{MSR-VTT}&HMDB51&UCF101\\
			\multirow{1}*{Method}&R@1&R@5&R@10&Top-1&Top-1\\
			\hline
			w/o MCQ&22.3&43.8&52.0&33.2&45.7\\
			Answer Random&23.0&45.5&55.5&36.9&50.7\\
			Answer Noun&24.9&46.2&\textbf{58.0}&36.2&51.8\\
			Answer Verb&23.3&\textbf{46.7}&57.5&36.3&51.5\\
			MWP&20.6&39.7&50.1&29.0&38.7\\
			Highest-level&23.3&46.0&56.4&36.5&47.7\\
			\multirow{1}*{Ours}&\textbf{26.0}&46.4&56.4&\textbf{37.7}&\textbf{53.1}\\
			\bottomrule[1pt]
	\end{tabular}}
	\vspace{-10pt}
	\label{tab:ablation}
\end{table}
\subsubsection{CLIP-based Pre-training }
Because of the prominent success of the CLIP~\cite{clip} (Contrastive Language-Image Pre-training) in learning image-text representations, which is pre-trained on 400 million image-text pairs, some recent work~\cite{straight,clip4clip} utilize the pre-trained CLIP for text-to-video retrieval. We also initialize our model from CLIP weights to pre-train a model following the setting of CLIP4Clip~\cite{clip4clip}. Specifically, we use the pre-trained CLIP (ViT-B/32) as the backbone of VideoFormer and TextFormer, and randomly initialize BridgeFormer. The comparisons between our method and other CLIP-initialized methods are shown in Table.~\ref{tab:clip}. We can observe that our CLIP-based pre-trained model achieves higher performance for text-to-video retrieval on three datasets with under both the zero-shot and fine-tune evaluation. Our pretext task MCQ also benefits CLIP-based video-text pre-training for downstream text-to-video retrieval. 

\begin{table}\centering
	\vspace{5pt}
	\caption{Ablation study on the effects of video information when answering the questions. Results on WebVid-2M validation set for noun or verb questions are reported.} 
	\vspace{-5pt} 
	\scalebox{0.8}{
		\begin{tabular}{c|ccc|ccc}
			\toprule[1pt]
			&\multicolumn{3}{c|}{Answer Noun}&\multicolumn{3}{c}{Answer Verb}\\
			\multirow{1}*{}&R@1&R@5&R@10&R@1&R@5&R@10\\
			\hline
			w/o Video&6.6&17.5&24.3&4.5&12.3&17.7\\
			with Video&58.6&81.1&87.2&40.7&64.0&73.2\\
			\bottomrule[1pt]
	\end{tabular}}
	\vspace{-10pt}
	\label{tab:video}
\end{table}

\subsection{Ablation Studies}
In this section, we discuss the effectiveness of our design on the pretext task MCQ through evaluating different models for zero-shot text-to-video retrieval on MSR-VTT, and zero-shot action recognition on HMDB51 and UCF101.

{\flushleft \bf Is MCQ effective?} Yes. As shown in Table.~\ref{tab:ablation}, pre-training a model without MCQ pretext task drops performance significantly, where only two separate encoders are adopted to contrast video-level and sentence-level features. 

{\flushleft \bf Does it help to answer noun and verb questions?} Yes.  As shown in Table.~\ref{tab:ablation}, training the BridgeFormer through answering noun questions only or verb questions only both harm performance. Randomly erasing words to construct questions also achieves worse results. 

{\flushleft \bf Do videos help to answer questions?} Yes. As shown in Table.~\ref{tab:video}, when the noun-question and verb-question select answers only through calculating the similarity between question representations and phrase representations from text encoder  without resorting to video tokens through BridgeFormer, the results decrease sharply. 

{\flushleft \bf Multiple Choice Questions \textit{vs.} Masked Word Prediction.} Training the BridgeFormer to predict the answer in the form of word tokens (similar to existing masked work prediction (MWP)) rather than select the correct answer in a batch of phrases in our MCQ actually hurts performance as shown in Table.~\ref{tab:ablation}, which is even lower than the baseline (w/o MCQ).

{\flushleft \bf All-level features \textit{vs.} highest-level features for BridgeFormer.} When BridgeFormer takes the highest-level features from the text and video encoders as inputs (a cascading structure) instead of all-level features (a parallel structure), we observe the performance drops as shown in Table.~\ref{tab:ablation} due to the lack of regularization on intermediate features. 
Even so, using only the highest-level features can also slightly outperform our baseline (w/o MCQ), indicating the effectiveness of our MCQ pretext task.
Actually, such a cascading structure is similar to those used in previous works \cite{clipbert,videobert,univl} where two separate encoders followed by a cross transformer are adopted. However, the cross transformer in these works cannot be easily removed in the same way as our BridgeFormer for downstream retrieval, \eg, evident 6.7\% decreases were observed in \cite{univl} in terms of R@1 on text-to-video retrieval, further indicating the flexibility and feasibility of our novel MCQ.

\begin{table}\centering\small
	\caption{The effects of the prompt ``[MASK]'' for noun and verb representations, where ``End'', ``Middle'' and ``Start'' denote the location of the prompt. For zero-shot text-to-video retrieval on MSR-VTT, \textbf{higher} R@k indicates better performance. For zero-shot action recognition on HMDB51 and UCF101, \textbf{higher} top-1 accuracy is better.} 
	\vspace{-5pt} 
	\scalebox{0.8}{
		\begin{tabular}{c|ccc|cc}
			\toprule[1pt]
			&\multicolumn{3}{c|}{MSR-VTT}&HMDB51&UCF101\\
			\multirow{1}*{Method}&R@1&R@5&R@10&Top-1&Top-1\\
			\hline
			w/o Prompt&23.1&43.5&54.3&34.8&45.8\\
			End&24.2&\textbf{45.7}&54.4&33.4&48.5\\
			Middle&24.3&43.2&53.9&33.1&46.4\\
			Start&\textbf{25.1}&45.4&\textbf{55.4}&\textbf{34.9}&\textbf{51.4}\\
			
			\bottomrule[1pt]
	\end{tabular}}
	\vspace{-10pt}
	\label{tab:prompt}
\end{table}

{\flushleft \bf Prompt for Phrase Representation} In our method, BridgeFormer is trained to select the correct answer by contrasting noun answer representations with noun representations, and contrasting verb answer representations with verb representations. Accurate representations for noun and verb phrases are essential. Since TextFormer is trained with full sentences, it fails to encode accurate representations for phrases when it takes a single noun or verb phrase as the input due to the lack of context. Motivated by the success of prompt engineering~\cite{clip}, we add ``[MASK]'' before the noun and verb phrase (\eg ``[MASK] [MASK] [MASK] green grass'')  to extract noun  or verb representations from TextFormer. We show ablation studies of the prompt ``[MASK]'' for noun and verb representations in Table.~\ref{tab:prompt}, where each model is pre-trained using 1 frame. The model without the prompt ``[MASK]'' takes a single noun or verb phrase as inputs, and achieves the worse results on both the zero-shot text-to-video retrieval and action recognition, showing that TextFormer cannot understand the semantics accurately with a single noun or verb phrase as inputs. The model with the prompt ``[MASK]'' at the beginning of the phrase achieves the best results in general, and we adopt this practice in our method.

\begin{table}\centering
	\vspace{-2pt}
	\caption{Comparisons between the video encoder in our method and Frozen~\cite{frozen}. The evaluation is performed on zero-shot text-to-video retrieval on MSR-VTT, where \textbf{higher} R@k and \textbf{lower} MdR (Median Rank) indicate better performance. ``\# Params'' denotes the number of parameters of the video encoder (M: million).} 
	\vspace{-5pt} 
	\scalebox{0.8}{
		\begin{tabular}{c|cccc|c}
			\toprule[1pt]
			\multirow{1}*{Method}&R@1&R@5&R@10&MdR&\# Params\\
			\hline
			Frozen~\cite{frozen}&18.7&39.5&51.6&10.0&114M\\
			Ours&\textbf{22.3}&\textbf{43.8}&\textbf{52.0}&\textbf{9.0}&86M\\
			\bottomrule[1pt]
	\end{tabular}}
	\vspace{-15pt}
	\label{tab:vit}
	\vspace{-1pt}
\end{table}

{\flushleft \bf Comparison of Video Encoder with Frozen.} Frozen~\cite{frozen} also adopts ViT~\cite{vit} as the video encoder, and adds temporal attention blocks based on the spatial attention blocks of ViT to encode videos with variable-length sequences. As shown in Table.~\ref{tab:vit}, compared with Frozen, our VideoFormer decreases 28 million parameters. Furthermore, the model without the pretext task MCQ indeed takes the same pre-training approach as Frozen except for the video encoder, and achieves better results for zero-shot text-to-video retrieval on MSR-VTT~\cite{msr}, which proves the efficiency and effectiveness of our VideoFormer.

\begin{figure}
	\centering
	\includegraphics[width=1.0\linewidth]{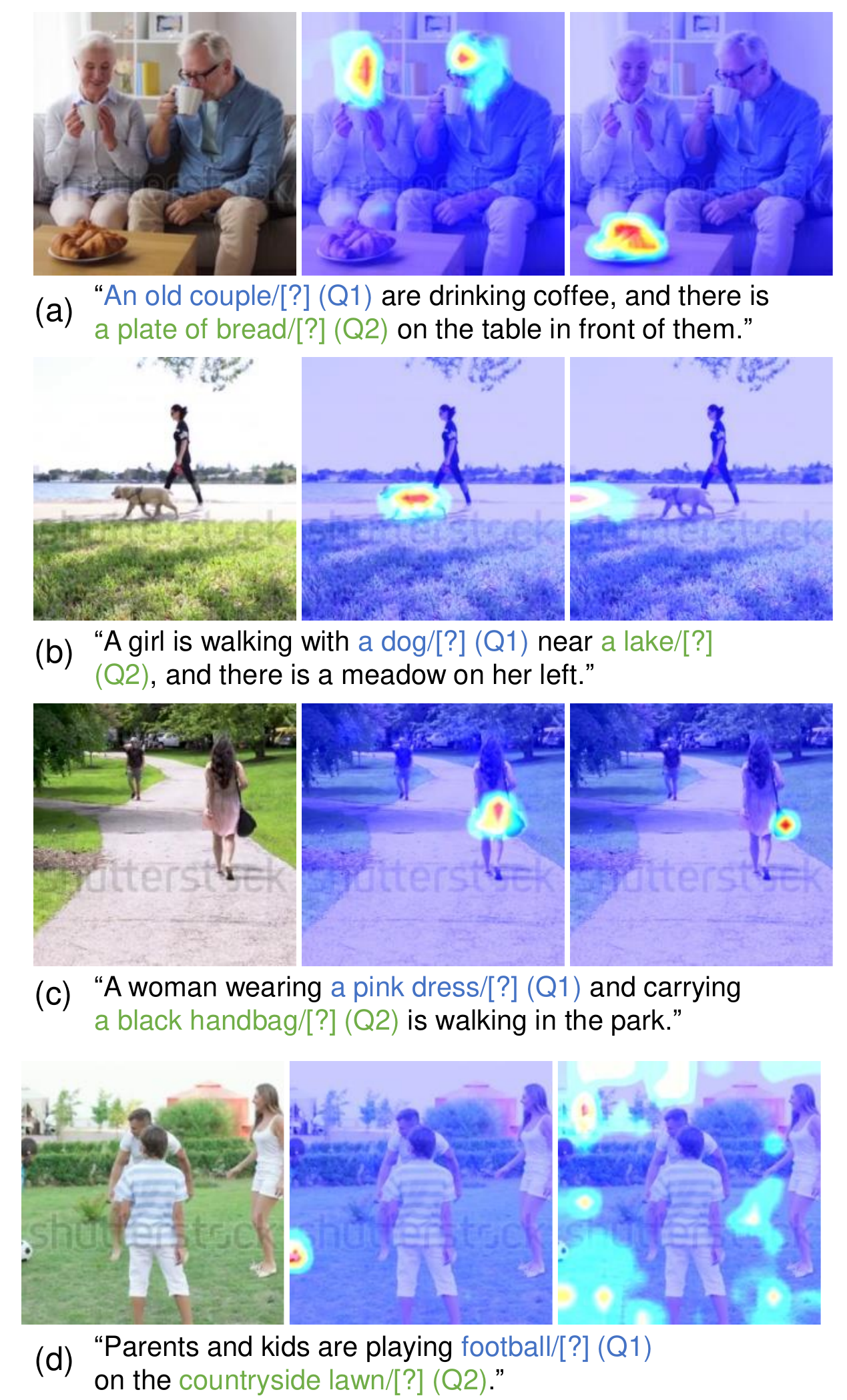}
	\vspace{-6mm}
	\caption{The visualization of the cross-modality attention between the text tokens of \textbf{noun questions} (as query) and video tokens (as key and value) from BridgeFormer. In the second column, the noun phrase marked in blue (Q1) is erased as the question, and in the third column, the noun phrase marked in green (Q2) is erased as the question. BridgeFormer attends to video patches with specific object information to answer noun questions.}
	\label{fig:vis}
	\vspace{-6mm}	
\end{figure}

\begin{figure}
	\centering
	\includegraphics[width=1.0\linewidth]{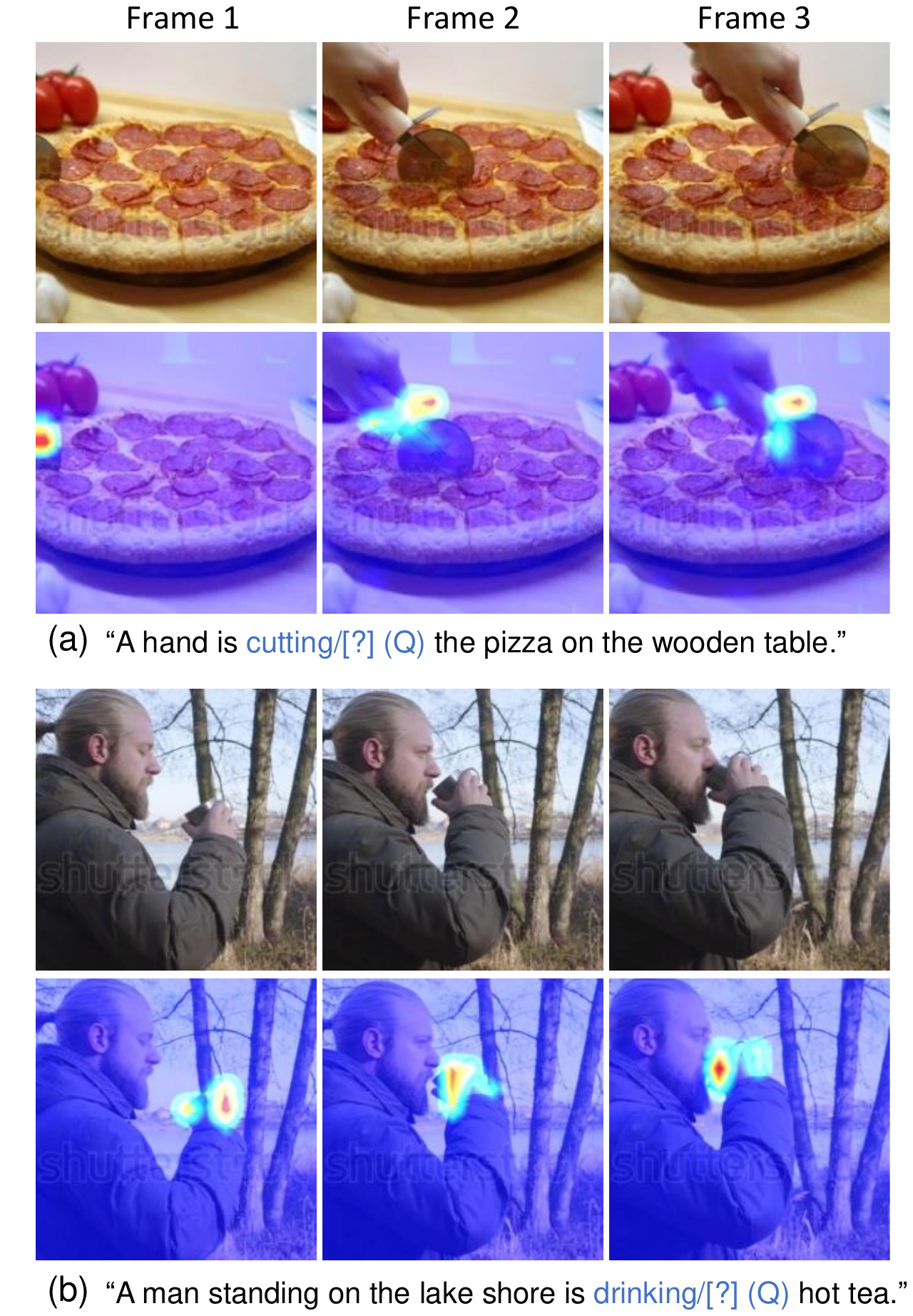}
	\vspace{-6mm}
	\caption{The visualization of the cross-modality attention between the text tokens of \textbf{verb questions} (as query) and video tokens (as key and value) from BridgeFormer. Three frames sampled from a video are shown and the verb phrase marked in blue (Q) is erased as the question. BridgeFormer focuses on object motions of video tokens to answer verb questions.}
	\label{fig:verb}
	\vspace{0mm}	
\end{figure}

\subsection{Visualization}
In our method, the pretext task MCQ is performed using a parametric module BridgeFormer, to answer multiple choice questions. We construct questions through erasing the content phrases (\ie noun and verb phrases) of the text, and BridgeFormer is trained to select the correct answer from multiple choices by resorting to the local tokens of VideoFormer. Specifically, given question text tokens from TextFormer as the query, and video tokens from VideoFormer as the key and value, BridgeFormer performs cross-modality attention between them. 

\subsubsection{Answering Noun Questions}
We first visualize the cross-modality attention between noun question tokens and video tokens in Fig.~\ref{fig:vis}. In the second column, the noun phrase marked in blue (Q1) is erased as the question, and in the third column, the noun phrase marked in green (Q2) is erased as the question. In Fig.~\ref{fig:vis} (a), when ``an old  couple'' is erased as the question (Q1), BridgeFormer focuses on video tokens that depict the appearance characteristics of the persons, and when ``a plate of bread'' is erased (Q2), it focuses on object video tokens on the table. In Fig.~\ref{fig:vis} (d), when ``football'' is erased (Q1), BridgeFormer focuses on the object video tokens that can be associated with ``play'', and when the location phrase ``countryside lawn'' is erased (Q2), it pays more attention to the video tokens in the background to infer the answer. BridgeFormer attends to video patches with specific object information to answer questions, which also shows that VideoFormer extracts accurate spatial content from videos. 

\subsubsection{Answering Verb Questions}
We further visualize the cross-modality attention between verb question tokens and video tokens in Fig.~\ref{fig:verb}. Three frames are sampled from a video and the verb phrase marked in blue is erased as the question. In Fig.~\ref{fig:verb} (a), when the verb ``cutting'' is erased, BridgeFormer focuses on the motion of the spoon on the pizza, and in Fig.~\ref{fig:verb} (b), when the verb ``drinking'' is erased, it follows the movement of the hand holding a cup of water. BridgeFormer focuses on object motions of video tokens to answer verb questions, which also shows that VideoFormer captures temporal dynamics of videos.

\section{Conclusion}
In this work, we introduce a novel pretext task, Multiple Choice Questions (MCQ) for video-text pre-training, which strengthens fine-grained semantic associations between local video and text features, and at the same time preserves high efficiency for retrieval. A parametric module BridgeFormer is trained to answer questions constructed by text features via resorting to video features, and can be readily removed for downstream tasks. Extensive evaluations on the text-to-video retrieval and zero-shot action recognition clearly show the great superiority of our method. 
{\flushleft \bf Limitation.} (1) Off-the-shelf NLP models can not  extract completely accurate noun and verb phrases for us to construct questions. (2) The text descriptions and corresponding videos may be actually misaligned in existing video-text datasets, leading to noisy supervision.
{\flushleft \bf Negative Social Impacts.} Since we do not filter out possible inappropriate videos (\eg, of blood and violence) in the pre-training dataset, our model can be used to search terrible videos for spreading. Utilizing the pre-trained model to filter out those videos and re-training a model can help.
{\flushleft \bf Acknowledgment} Ping Luo is supported by the General Research Fund of HK No.27208720 and 17212120.

\bibliographystyle{ieee_fullname}
\bibliography{egbib.bib}

\begin{thebibliography}{10}\itemsep=-1pt

\bibitem{vatt}
Hassan Akbari, Linagzhe Yuan, Rui Qian, Wei-Hong Chuang, Shih-Fu Chang, Yin
  Cui, and Boqing Gong.
\newblock Vatt: Transformers for multimodal self-supervised learning from raw
  video, audio and text.
\newblock {\em arXiv preprint arXiv:2104.11178}, 2021.

\bibitem{MMV}
Jean-Baptiste Alayrac, Adria Recasens, Rosalia Schneider, Relja Arandjelovic,
  Jason Ramapuram, Jeffrey De~Fauw, Lucas Smaira, Sander Dieleman, and Andrew
  Zisserman.
\newblock Self-supervised multimodal versatile networks.
\newblock {\em NeurIPS}, 2(6):7, 2020.

\bibitem{XDC}
Humam Alwassel, Dhruv Mahajan, Bruno Korbar, Lorenzo Torresani, Bernard Ghanem,
  and Du Tran.
\newblock Self-supervised learning by cross-modal audio-video clustering.
\newblock In {\em NeurIPS 2020}. NeurIPS, 2020.

\bibitem{noise}
Elad Amrani, Rami Ben-Ari, Daniel Rotman, and Alex Bronstein.
\newblock Noise estimation using density estimation for self-supervised
  multimodal learning.
\newblock In {\em AAAI}, volume~35, pages 6644--6652, 2021.

\bibitem{didemo}
Lisa Anne~Hendricks, Oliver Wang, Eli Shechtman, Josef Sivic, Trevor Darrell,
  and Bryan Russell.
\newblock Localizing moments in video with natural language.
\newblock In {\em ICCV}, pages 5803--5812, 2017.

\bibitem{frozen}
Max Bain, Arsha Nagrani, G{\"u}l Varol, and Andrew Zisserman.
\newblock Frozen in time: A joint video and image encoder for end-to-end
  retrieval, 2021.

\bibitem{qa2}
Aman Chadha, Gurneet Arora, and Navpreet Kaloty.
\newblock iperceive: Applying common-sense reasoning to multi-modal dense video
  captioning and video question answering.
\newblock {\em arXiv preprint arXiv:2011.07735}, 2020.

\bibitem{mcn}
Brian Chen, Andrew Rouditchenko, Kevin Duarte, Hilde Kuehne, Samuel Thomas,
  Angie Boggust, Rameswar Panda, Brian Kingsbury, Rogerio Feris, David Harwath,
  et~al.
\newblock Multimodal clustering networks for self-supervised learning from
  unlabeled videos.
\newblock {\em arXiv preprint arXiv:2104.12671}, 2021.

\bibitem{msvd}
David Chen and William~B Dolan.
\newblock Collecting highly parallel data for paraphrase evaluation.
\newblock In {\em ACL}, pages 190--200, 2011.

\bibitem{re1}
Shizhe Chen, Yida Zhao, Qin Jin, and Qi Wu.
\newblock Fine-grained video-text retrieval with hierarchical graph reasoning.
\newblock In {\em CVPR}, pages 10638--10647, 2020.

\bibitem{bert}
Jacob Devlin, Ming-Wei Chang, Kenton Lee, and Kristina Toutanova.
\newblock Bert: Pre-training of deep bidirectional transformers for language
  understanding.
\newblock In {\em NAACL}, 2019.

\bibitem{vit}
Alexey Dosovitskiy, Lucas Beyer, Alexander Kolesnikov, Dirk Weissenborn,
  Xiaohua Zhai, Thomas Unterthiner, Mostafa Dehghani, Matthias Minderer, Georg
  Heigold, Sylvain Gelly, et~al.
\newblock An image is worth 16x16 words: Transformers for image recognition at
  scale.
\newblock In {\em ICLR}, 2020.

\bibitem{multi}
Valentin Gabeur, Chen Sun, Karteek Alahari, and Cordelia Schmid.
\newblock Multi-modal transformer for video retrieval.
\newblock In {\em ECCV}, pages 214--229, 2020.

\bibitem{ig65}
Deepti Ghadiyaram, Du Tran, and Dhruv Mahajan.
\newblock Large-scale weakly-supervised pre-training for video action
  recognition.
\newblock In {\em CVPR}, pages 12046--12055, 2019.

\bibitem{coot}
Simon Ging, Mohammadreza Zolfaghari, H Pirsiavash, and Thomas Brox.
\newblock Coot: Cooperative hierarchical transformer for video-text
  representation learning.
\newblock In {\em NeurIPS}, 2020.

\bibitem{MemDPC}
Tengda Han, Weidi Xie, and Andrew Zisserman.
\newblock Memory-augmented dense predictive coding for video representation
  learning.
\newblock In {\em ECCV}, pages 312--329, 2020.

\bibitem{COCLR}
Tengda Han, Weidi Xie, and Andrew Zisserman.
\newblock Self-supervised co-training for video representation learning.
\newblock {\em NeurIPS}, 33:5679--5690, 2020.

\bibitem{contrastive2}
Rafal Jozefowicz, Oriol Vinyals, Mike Schuster, Noam Shazeer, and Yonghui Wu.
\newblock Exploring the limits of language modeling.
\newblock {\em arXiv preprint arXiv:1602.02410}, 2016.

\bibitem{vilt}
Wonjae Kim, Bokyung Son, and Ildoo Kim.
\newblock Vilt: Vision-and-language transformer without convolution or region
  supervision.
\newblock {\em arXiv preprint arXiv:2102.03334}, 2021.

\bibitem{CCL}
Quan Kong, Wenpeng Wei, Ziwei Deng, Tomoaki Yoshinaga, and Tomokazu Murakami.
\newblock Cycle-contrast for self-supervised video representation learning.
\newblock 2020.

\bibitem{hmdb}
Hildegard Kuehne, Hueihan Jhuang, Est{\'\i}baliz Garrote, Tomaso Poggio, and
  Thomas Serre.
\newblock Hmdb: a large video database for human motion recognition.
\newblock In {\em ICCV}, pages 2556--2563. IEEE, 2011.

\bibitem{clipbert}
Jie Lei, Linjie Li, Luowei Zhou, Zhe Gan, Tamara~L Berg, Mohit Bansal, and
  Jingjing Liu.
\newblock Less is more: Clipbert for video-and-language learning via sparse
  sampling.
\newblock In {\em CVPR}, pages 7331--7341, 2021.

\bibitem{hero}
Linjie Li, Yen-Chun Chen, Yu Cheng, Zhe Gan, Licheng Yu, and Jingjing Liu.
\newblock Hero: Hierarchical encoder for video+ language omni-representation
  pre-training.
\newblock In {\em EMNLP}, pages 2046--2065, 2020.

\bibitem{expert}
Yang Liu, Samuel Albanie, Arsha Nagrani, and Andrew Zisserman.
\newblock Use what you have: Video retrieval using representations from
  collaborative experts.
\newblock {\em arXiv preprint arXiv:1907.13487}, 2019.

\bibitem{univl}
Huaishao Luo, Lei Ji, Botian Shi, Haoyang Huang, Nan Duan, Tianrui Li, Jason
  Li, Taroon Bharti, and Ming Zhou.
\newblock Univl: A unified video and language pre-training model for multimodal
  understanding and generation.
\newblock {\em arXiv preprint arXiv:2002.06353}, 2020.

\bibitem{clip4clip}
Huaishao Luo, Lei Ji, Ming Zhong, Yang Chen, Wen Lei, Nan Duan, and Tianrui Li.
\newblock Clip4clip: An empirical study of clip for end to end video clip
  retrieval.
\newblock {\em arXiv preprint arXiv:2104.08860}, 2021.

\bibitem{qa4}
Tegan Maharaj, Nicolas Ballas, Anna Rohrbach, Aaron Courville, and Christopher
  Pal.
\newblock A dataset and exploration of models for understanding video data
  through fill-in-the-blank question-answering.
\newblock In {\em CVPR}, pages 6884--6893, 2017.

\bibitem{MIL-NCE}
Antoine Miech, Jean-Baptiste Alayrac, Lucas Smaira, Ivan Laptev, Josef Sivic,
  and Andrew Zisserman.
\newblock End-to-end learning of visual representations from uncurated
  instructional videos.
\newblock In {\em CVPR}, pages 9879--9889, 2020.

\bibitem{howto100m}
Antoine Miech, Dimitri Zhukov, Jean-Baptiste Alayrac, Makarand Tapaswi, Ivan
  Laptev, and Josef Sivic.
\newblock Howto100m: Learning a text-video embedding by watching hundred
  million narrated video clips.
\newblock In {\em ICCV}, pages 2630--2640, 2019.

\bibitem{contrastive1}
Aaron van~den Oord, Yazhe Li, and Oriol Vinyals.
\newblock Representation learning with contrastive predictive coding.
\newblock {\em arXiv preprint arXiv:1807.03748}, 2018.

\bibitem{support}
Mandela Patrick, Po-Yao Huang, Yuki Asano, Florian Metze, Alexander~G
  Hauptmann, Joao~F Henriques, and Andrea Vedaldi.
\newblock Support-set bottlenecks for video-text representation learning.
\newblock In {\em ICLR}, 2020.

\bibitem{elo}
AJ Piergiovanni, Anelia Angelova, and Michael~S Ryoo.
\newblock Evolving losses for unsupervised video representation learning.
\newblock In {\em CVPR}, pages 133--142, 2020.

\bibitem{straight}
Jes{\'u}s~Andr{\'e}s Portillo-Quintero, Jos{\'e}~Carlos Ortiz-Bayliss, and Hugo
  Terashima-Mar{\'\i}n.
\newblock A straightforward framework for video retrieval using clip.
\newblock In {\em Mexican Conference on Pattern Recognition}, pages 3--12.
  Springer, 2021.

\bibitem{clip}
Alec Radford, Jong~Wook Kim, Chris Hallacy, Aditya Ramesh, Gabriel Goh,
  Sandhini Agarwal, Girish Sastry, Amanda Askell, Pamela Mishkin, Jack Clark,
  et~al.
\newblock Learning transferable visual models from natural language
  supervision.
\newblock {\em arXiv preprint arXiv:2103.00020}, 2021.

\bibitem{lsmdc}
Anna Rohrbach, Marcus Rohrbach, Niket Tandon, and Bernt Schiele.
\newblock A dataset for movie description.
\newblock In {\em CVPR}, pages 3202--3212, 2015.

\bibitem{avlnet}
Andrew Rouditchenko, Angie Boggust, David Harwath, Brian Chen, Dhiraj Joshi,
  Samuel Thomas, Kartik Audhkhasi, Hilde Kuehne, Rameswar Panda, Rogerio Feris,
  et~al.
\newblock Avlnet: Learning audio-visual language representations from
  instructional videos.
\newblock {\em arXiv preprint arXiv:2006.09199}, 2020.

\bibitem{qa3}
Arka Sadhu, Kan Chen, and Ram Nevatia.
\newblock Video question answering with phrases via semantic roles.
\newblock {\em arXiv preprint arXiv:2104.03762}, 2021.

\bibitem{distilbert}
Victor Sanh, Lysandre Debut, Julien Chaumond, and Thomas Wolf.
\newblock Distilbert, a distilled version of bert: smaller, faster, cheaper and
  lighter.
\newblock {\em arXiv preprint arXiv:1910.01108}, 2019.

\bibitem{cc3m}
Piyush Sharma, Nan Ding, Sebastian Goodman, and Radu Soricut.
\newblock Conceptual captions: A cleaned, hypernymed, image alt-text dataset
  for automatic image captioning.
\newblock In {\em ACL}, pages 2556--2565, 2018.

\bibitem{ucf}
Khurram Soomro, Amir~Roshan Zamir, and Mubarak Shah.
\newblock Ucf101: A dataset of 101 human actions classes from videos in the
  wild.
\newblock {\em arXiv preprint arXiv:1212.0402}, 2012.

\bibitem{CBT}
Chen Sun, Fabien Baradel, Kevin Murphy, and Cordelia Schmid.
\newblock Learning video representations using contrastive bidirectional
  transformer.
\newblock {\em arXiv preprint arXiv:1906.05743}, 2019.

\bibitem{videobert}
Chen Sun, Austin Myers, Carl Vondrick, Kevin Murphy, and Cordelia Schmid.
\newblock Videobert: A joint model for video and language representation
  learning.
\newblock In {\em ICCV}, pages 7464--7473, 2019.

\bibitem{3d}
Du Tran, Heng Wang, Lorenzo Torresani, Jamie Ray, Yann LeCun, and Manohar
  Paluri.
\newblock A closer look at spatiotemporal convolutions for action recognition.
\newblock In {\em CVPR}, pages 6450--6459, 2018.

\bibitem{re2}
Michael Wray, Diane Larlus, Gabriela Csurka, and Dima Damen.
\newblock Fine-grained action retrieval through multiple parts-of-speech
  embeddings.
\newblock In {\em ICCV}, pages 450--459, 2019.

\bibitem{vlm}
Hu Xu, Gargi Ghosh, Po-Yao Huang, Prahal Arora, Masoumeh Aminzadeh, Christoph
  Feichtenhofer, Florian Metze, and Luke Zettlemoyer.
\newblock Vlm: Task-agnostic video-language model pre-training for video
  understanding.
\newblock {\em arXiv preprint arXiv:2105.09996}, 2021.

\bibitem{videoclip}
Hu Xu, Gargi Ghosh, Po-Yao Huang, Dmytro Okhonko, Armen Aghajanyan, Florian
  Metze, Luke Zettlemoyer, and Christoph Feichtenhofer.
\newblock Videoclip: Contrastive pre-training for zero-shot video-text
  understanding.
\newblock {\em arXiv preprint arXiv:2109.14084}, 2021.

\bibitem{msr}
Jun Xu, Tao Mei, Ting Yao, and Yong Rui.
\newblock Msr-vtt: A large video description dataset for bridging video and
  language.
\newblock In {\em CVPR}, pages 5288--5296, 2016.

\bibitem{re3}
Ran Xu, Caiming Xiong, Wei Chen, and Jason Corso.
\newblock Jointly modeling deep video and compositional text to bridge vision
  and language in a unified framework.
\newblock In {\em AAAI}, volume~29, 2015.

\bibitem{qa1}
Antoine Yang, Antoine Miech, Josef Sivic, Ivan Laptev, and Cordelia Schmid.
\newblock Just ask: Learning to answer questions from millions of narrated
  videos.
\newblock In {\em ICCV}, pages 1686--1697, 2021.

\bibitem{taco}
Jianwei Yang, Yonatan Bisk, and Jianfeng Gao.
\newblock Taco: Token-aware cascade contrastive learning for video-text
  alignment.
\newblock In {\em ICCV}, pages 11562--11572, 2021.

\bibitem{actbert}
Linchao Zhu and Yi Yang.
\newblock Actbert: Learning global-local video-text representations.
\newblock In {\em CVPR}, pages 8746--8755, 2020.

\bibitem{re4}
Dimitri Zhukov, Jean-Baptiste Alayrac, Ramazan~Gokberk Cinbis, David Fouhey,
  Ivan Laptev, and Josef Sivic.
\newblock Cross-task weakly supervised learning from instructional videos.
\newblock In {\em CVPR}, pages 3537--3545, 2019.

\end{thebibliography}

\end{document}